\documentclass[letterpaper, 10 pt, conference]{ieeeconf}

\IEEEoverridecommandlockouts                              % This command is only needed if 
\usepackage{graphics} % for pdf, bitmapped graphics files
\usepackage{graphicx}
\usepackage{mathptmx} % assumes new font selection scheme installed
\usepackage{times} % assumes new font selection scheme installed
\usepackage{amsmath} % assumes amsmath package installed
\usepackage{amssymb}  % assumes amsmath package installed\
\usepackage{subcaption}
\usepackage{url}
\usepackage[colorinlistoftodos, textwidth=4cm, shadow]{todonotes}
\usepackage{flushend}
\usepackage{siunitx}
\title{\LARGE \bf
Phytoplankton Hotspot Prediction With an Unsupervised Spatial Community Model
}

\author{Arnold Kalmbach$^{1}$, Yogesh Girdhar$^{2}$, Heidi M. Sosik$^{3}$ and Gregory Dudek$^{1}$% <-this % stops a space
\thanks{$^{1}$A. Kalmbach and G. Dudek are with the Centre for Intelligent Machines, School of Computer Science, McGill University, 3480 University St. Montr\'{e}al, QC, Canada
        {\tt\small \{akalmbach,dudek\}@cim.mcgill.ca}}%
\thanks{$^{2}$Y. Girdhar is with the Woods Hole Oceanographic Institution, Applied Ocean Physics and Engineering Department, Woods Hole, MA 02543
        {\tt\small yogi@whoi.edu}}%
\thanks{$^{3}$H.M. Sosik is with the Woods Hole Oceanographic Institution, Biology Department, Woods Hole, MA 02543
        {\tt\small hsosik@whoi.edu}}%
}

\begin{document}

\maketitle
\thispagestyle{empty}
\pagestyle{empty}

%%%%%%%%%%%%%%%%%%%%%%%%%%%%%%%%%%%%%%%%%%%%%%%%%%%%%%%%%%%%%%%%%%%%%%%%%%%%%%%%
\begin{abstract}
Many interesting natural phenomena are sparsely distributed and discrete. Locating the hotspots of such sparsely distributed
phenomena is often difficult because their density gradient is likely to be very noisy. We present a novel approach to this search problem, where we model the co-occurrence relations between a robot's observations with a Bayesian nonparametric topic model. This approach makes it possible to produce a robust estimate of the spatial distribution of the target, even in the absence of direct target observations. We apply the proposed approach to the problem of finding the spatial locations of the hotspots of a specific phytoplankton taxon in the ocean. We use classified image data from Imaging FlowCytobot (IFCB), which automatically measures individual microscopic cells and colonies of cells. Given these individual taxon-specific observations, we learn a phytoplankton community model that characterizes the co-occurrence relations between taxa. We present experiments with simulated robot missions drawn from real observation data collected during a research cruise traversing the US Atlantic coast. Our results show that the proposed approach outperforms nearest neighbor and k-means based methods for predicting the spatial distribution of hotspots from in-situ observations. 

\end{abstract}

%%%%%%%%%%%%%%%%%%%%%%%%%%%%%%%%%%%%%%%%%%%%%%%%%%%%%%%%%%%%%%%%%%%%%%%%%%%%%%%%
\section{Introduction}
This paper addresses the problem of finding spatial density hotspots of a sparsely distributed target phenomenon. We hypothesize that by modeling distributions of co-occurring phenomena, we can predict the presence of the target phenomenon, even in the absence of its direct observation.  In particular, we focus on the problem of finding hotspots of target phytoplankton taxa in in-situ observations made by a robotic marine instrument following a fixed survey trajectory.

\begin{figure}
\includegraphics[width=\linewidth]{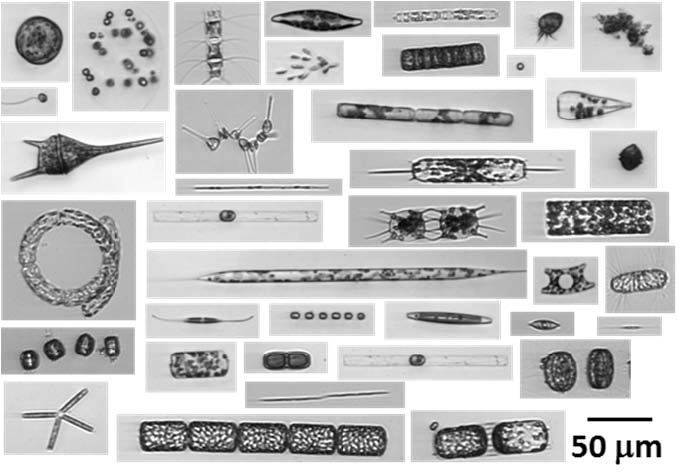}
\caption{Example of  images captured by the Imaging FlowCytobot (IFCB). These images are classified into 47 classes corresponding to various phytoplankton taxa and other particle types (e.g., detritus) \cite{Olson2007a}. The proposed topic model automatically discovers community structure from the taxon-specific observational data. We use the model to predict likelihood of observing a target taxon in a given location, without the need for any direct observations of the target.}
\label{fig:IFCB}
\end{figure}

Phytoplankton are microscopic organisms that form the base of marine food webs.
They produce chlorophyll and other pigments to harvest sunlight and fuel photosynthesis, so they can utilize $\mathrm{CO}_2$ and other nutrients to produce $\mathrm{O}_2$ and new organic matter. As such, they play critical roles in global biogeochemical cycles and in structuring marine ecosystems.  
Marine scientists have long used techniques to measure the amount of chlorophyll in a water sample as a proxy for phytoplankton biomass \cite{Lorenzen1966}. These methods are coarse and give only bulk indices, with no information about which species of phytoplankton are present. Phytoplankton are extremely diverse, however, and their community structure plays a major role in shaping ecosystems and their functions. As an extreme example, particular species are known to cause toxic blooms that can threaten wildlife as well as human health.

To meet the gap in observational capability that includes taxonomic resolution, Sosik and Olson have developed the automated, submersible Imaging FlowCytobot (IFCB)~\cite{Olson2007a} and a coupled analysis system \cite{Sosik2007,Sosik2016}. This system can  detect and classify phytoplankton automatically in small samples of ocean water collected serially over long periods of time (weeks to years). The images acquired by IFCB have high enough resolution ($\sim$\SI{1}{\micro\meter}) that many can be classified to genus or species (Fig. \ref{fig:IFCB}). Currently, the IFCB can be routinely moored in the ocean or continuously sample underway on a ship. In addition, prototype deployments have demonstrated its capability to operate on robotic surface vehicles.

In this work we use the detections and detection locations produced by the IFCB as input to the proposed technique, which can enable a marine robot to detect hotpots of sparsely distributed plankton species. 

\subsection*{Contributions}
We present a novel way to robustly estimate the spatial density of a sparsely distributed natural phenomena -- phytoplankton taxa -- using  a probabilistic generative model. The observed distribution of plankton taxa at a location is modelled as a sparse mixture of communities, and the communities are modeled as sparse mixture of plankton taxa. In addition, the model puts smoothness constrains on the spatial distribution of communities. The proposed community model allows us to reason about which plankton taxa we expect to observe together in situations where not all species can be observed.

We demonstrate that our model is able to predict `hotspots' i.e., locations where a particular taxon obtains high probability of being observed -- based on the distribution of the other taxa in a survey. We compare our model's performance in this task to
two other strategies: (1) an exhaustive search representing the best any model can be expected to perform if the training and testing data are drawn from the same distribution, but which has a higher computational complexity of than our approach; and (2) a k-means based strategy that has an equivalent complexity to our approach. We show that our model outperforms both other strategies when training and testing data are taken from separate parts of the world, and is competitive when training and testing data are near to one another.

\section{Related Work}
With recent improvements in in-situ sensing and adaptive sampling algorithms, robots are being used to detect and track many different kinds of natural phenomena underwater. For example, Zhang et al.~\cite{Zhang2016, Zhang2012} have demonstrated a technique to autonomously track upwelling fronts in space and time. Ocean upwelling refers to the processes by which nutrient-rich water from the deeper ocean is transported to the surface. Coastal upwelling zones are typically hot-spots for phytoplankton and zooplankton. The authors identify upwelling by detecting vertical temperature gradients.

Much recent work in robotic marine tracking has focused on using visual cues to enumerate species or other phenomena using adaptive sampling techniques~\cite{ShkurtiIRoS2012, manjanna2016efficient, ECE3:ECE32701}. Typical vision systems are much too coarse to provide measurements of phytoplankton populations. The present work makes use of the novel vision capabilities of the IFCB to move towards using similar approaches for phytoplankton tracking applications.

Chlorophyll fluorescence sensors provide a way to detect phytoplankton directly. For example, Das et al.~\cite{Das2012} used fluorometers on AUVs and Lagrangian drifters to locate and track phytoplankton patches in the ocean.

Das et al. \cite{Das2015} also developed an approach to predict the abundance of a particular species known to cause harmful algal blooms in the study region. Their objective was to optimize capture of the target species in a small, fixed number of physical samples taken by a robot. Their model is based on a Gaussian Process, with a set of environment variables including fluorescence, temperature, and other chemical properties as inputs, and the results of manual molecular analysis of historical data as training targets.
Whereas their method focuses on predicting the abundance of the target species from environmental variables, our method predicts the \textit{relative} abundance of a taxon from the distribution of other taxa. These two perspectives are complementary and both are useful for the problem of automatically choosing the best set of sample locations for extended ex-situ analysis.

Rao et al. \cite{Rao2016} proposed the use of a neural network to learn a shared representation over multiple sensor modalities for underwater vehicles (imagery and bathymetry). The learned model is then used to identify information-rich locations given exclusively the bathymetric data. For a small number of classes, this type of multimodal learning framework might capture more of the spatial or temporal complexities of plankton taxon associations. However, as the number of modalities increases this approach is not scalable and therefore it is not suitable for modelling the numerous plankton taxa we consider from this dataset.

Topic modeling \cite{Blei2012} offers a natural way to represent highly multimodal data such as the spatio-temporal distribution of plankton taxa. Topic models specify a generative model of the data, where each set of discrete observations is modeled as a mixture of topics (plankton communities) and, in turn, each topic or community is modeled as a mixture of plankton taxa. In topic models, Dirichlet distribution or Dirichlet process \cite{Teh2010} priors can be used to control the sparseness of the taxonomic distribution representing a community, and the sparseness of the community distribution at a given location. Girdhar et al.~\cite{Girdhar2013IJRR, Girdhar2016} extended the standard topic model to account for spatial and temporal correlation of observations. The plankton community model we propose here is based on the Bayesian nonparametric spatio-temporal topic model (BNP-ROST) \cite{Girdhar2016}.

\section{Approach}

We are interested in identifying areas of the ocean where we are most likely to observe a particular class of plankton. Let $w$ be a plankton observation, such that $w \in [1,V]$, where $V$ is the total number of known plankton taxa, and let $x$ be the spatio-temporal coordinates of this observation, i.e. the vector \emph{[Time since cruise start, Eastings, Northings]}, which we refer to simply as the \emph{location}. The goal is then to estimate the distribution $P(w=v|x, W, X)$, i.e., the distribution of classes $v$ at location $x$, given all previous observations $W$ and their locations $X$. We define a hotspot as the set of locations where the probability of observing a class exceeds a class-specific threshold. 

Given the high dimensionality of the distribution of classes, we propose the following factorization to approximate the target distribution.
\begin{equation}
    P(w=v|x, W, X) = \sum_k P(w=v|z=k) P(z=k | x).
\end{equation}

Here $z$ is a latent variable, which essentially denotes a plankton community, and the distribution $P(w=v|z=k)$ models the likelihood that the an observation is of taxon $v$ given that it was drawn from community $k$. The distribution $P(z=k | x)$ models the spatio-temporal distribution of community $k$. 

We model $P(w|z)$ with a Dirichlet prior. This assumption ensures that our model assigns higher probability to communities represented by sparse taxon distributions. The posterior distribution can be expressed in terms of observation counts:
\begin{equation}
    \Phi=P(w=v|z=k,\beta) = \frac{N^{(v)}_k + \beta}{N^{(.)}_k+V\beta},
\end{equation}
where $\beta$ is a symmetric Dirichlet distribution parameter. $\Phi=\{\phi_{v,k}\}$ is a $V \times K$ matrix that represents the community model.

We assume that the number of plankton communities is unknown and use a variant of the Chinese restaurant process (CRP) \cite{Teh:2006:HDP, Teh2010} to model the prior for the distribution of communities at a given location. The posterior community distribution at location $x$ is given by:  

\begin{eqnarray}
\Theta=P(z=k | x, \alpha, \gamma ) \propto  
 \begin{cases}
     {N^{k}_{g(x)} + \alpha} & k \in Z \\
     \gamma & k \notin Z
 \end{cases}
\end{eqnarray}

Here $Z$ is the current set of all known plankton communities, $N^{k}_{g(x)}$ is the number of times we have observed a member of community $k$ in the spatio-temporal neighborhood of location $x$, and $\alpha, \gamma$ are model hyperparameters. Hence, with high probability proportional to $N^{k}_{g(x)} + \alpha$, the observation belongs to community $k$ that is common around location $x$, and with a small probability proportional to $\gamma$, the observation belongs to a new, un-modeled community.

We divide the world into spatio-temporal cells such that the cell that contains location $x$ is denoted by $c(x)$, and then define $g(x)$ to be the set of cells in the Von Neumann neighborhood of $c(x)$. The spatio-temporal distribution of communities can then be modeled by $\Theta=\{\theta_{c,k}\}$, which is a $C \times K$ matrix, where $C$ is the total number of spatial cells in the world that have observations. 

When the robot explores a new location where the number of observations of the target taxon is zero or too small to be statistically significant, we hypothesize that the plankton topic model can be used to compute a robust estimate of the likelihood of observing the target taxon on the basis of its association with other taxa.

To accomplish this, first we learn the plankton community topic model from previously visited locations. This data could come either from previous missions or locations visited earlier on the same mission. We then compute the maximum likelihood topic assignments for the observations in the neighborhood of the target location. Finally, given the topic assignments and the original model, we can compute the maximum likelihood distribution for all classes, including an unobserved taxon in the target location:

\begin{equation}
    P(w=v|x,\Phi) = \sum_k \theta^*_{c(x),k} \phi_{v,k}
\end{equation}

Here $\theta^*$ is the maximum likelihood topic distribution in the neighborhood of target location $x$. 

To learn the community model, we use an online Gibbs sampler \cite{Girdhar2015Gibbs}, which equally divides the computational resources between computing the posterior topic distribution of the most recent observation, and updating the topic labels and the topic model over the previous observations. 

\section{Experiment}
To evaluate the hypothesis that the proposed plankton community model can be used to predict hotspots of a target class, we present experiments with simulated missions, drawn from real data, focusing on the worst case scenario where no observations of the target class have been made. 

\begin{figure}
\includegraphics[width=\linewidth]{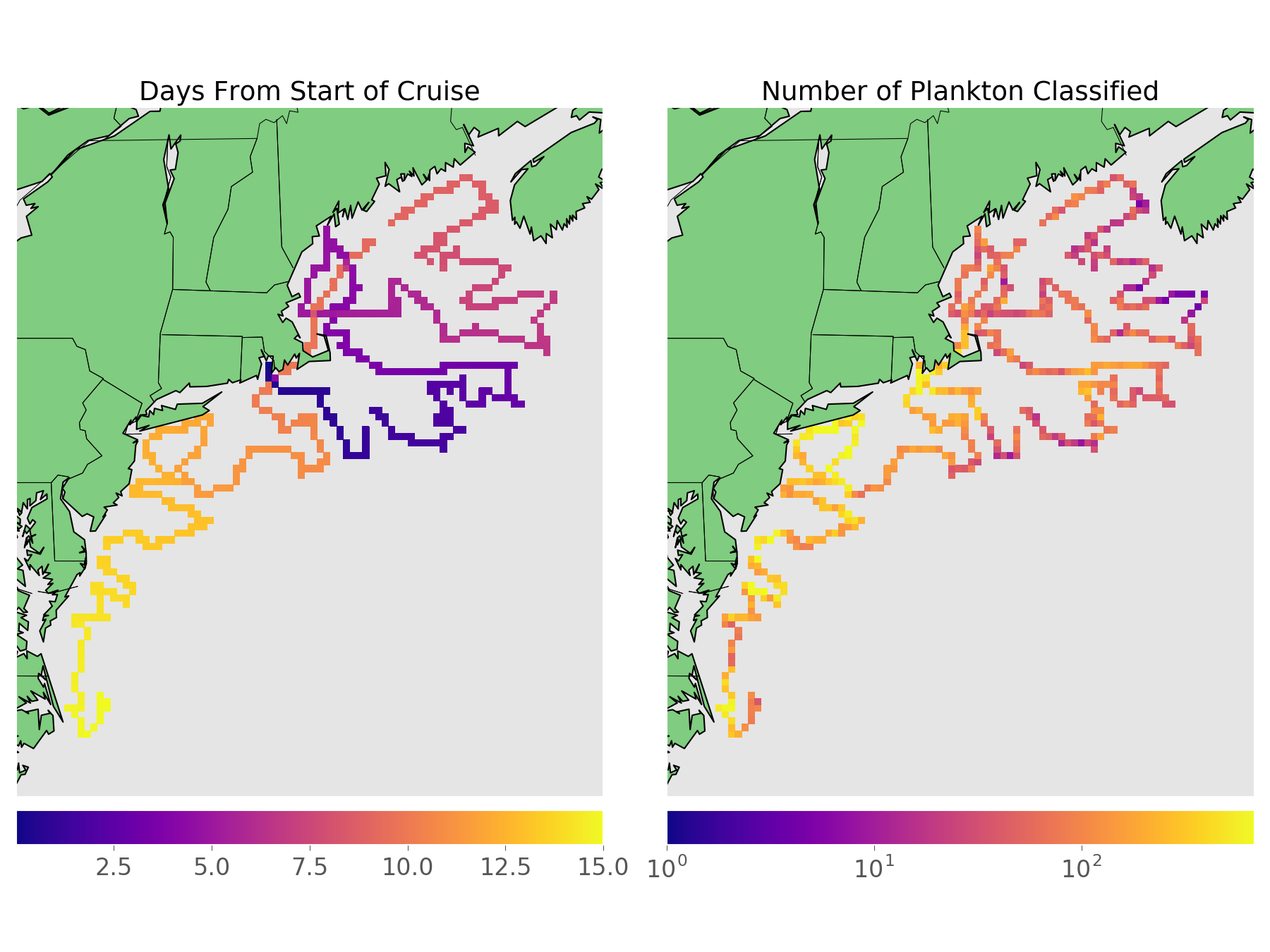}
\caption{Summary of data recorded during the Pisces 14-05 cruise. Left, color shows progress in time. Right, color shows the number of plankton observed at each sample location.}
\label{fig:summary_in}
\end{figure}

We validate our approach with IFCB classification results from NOAA's Fall 2014 EcoMon Survey aboard the Research Vessel Pisces (Cruise PC 14-05). The IFCB was configured to automatically sample from underway flowing surface seawater (5 ml approximately every 20 min) during the period 4-19 November 2014. The classification system generated over 140,000 individual phytoplankton observations from these water samples. Classification results comprise a dataset with 47 taxa at 852 locations spanning the US Atlantic coast from North Carolina to Maine (See Fig.~\ref{fig:summary_in}).\footnote{The dataset is available online at \url{http://ifcb-data.whoi.edu/IFCB102_PiscesNov2014}}

We divide the sample locations into equal-sized parts, representing the training and test phases of the simulated mission. The counts of all 47 taxa were kept in the training set and used to learn the topic model. For the test set, we held out each of the 8 most-frequently observed phytoplankton taxa one at a time. We define the hotspots of
a taxon to be the top 50 sample locations in the test data, where the relative abundance of the taxon to all other taxa was highest. The 8 tested taxa make up just over 81\% of all the observations in the dataset. The most common taxa are miscellaneous centric diatom chains (``mix\_elongated''), mixed species of pennate diatoms, \emph{Thalassiosira} spp., \emph{Guinardia delicatula, Guinardia striata, Dictyocha} spp., \emph{Ephemera} spp., and \emph{Phaeocystis} spp.

With the topic model learned from the training data we compute the maximum likelihood topic assignments for the test data. 
To simulate the case when there are no observations of the target taxon $v^\star$, we use $\Phi^{(\neg v^\star)}$ instead:
\begin{equation}
  \phi^{(\neg v^\star)}_{k,v} \triangleq p\left(w_i = v | z_i = k, \mathbf{z_{\neg v^\star}} \right) =  \frac{N^{(v)}_{-i,k} + \beta}{N^{(\neg v^\star)}_{-i,k} + \left(V-1\right)\beta}
\end{equation}

The maximum likelihood taxon distribution for the test data is given by $\Theta \Phi^{(\neg v^\star)}$, but since we do not update the topics given the new data, we can instead estimate $P(w = v^\star|c) = \sum_k \theta_{c,k} \phi_{k,v^\star}$. While our method accounts for the sparsity of taxon distributions, this dataset also features
sparsity in terms of the locations of observations. To address this separate issue, we resort to a 2D spatial median filter. Finally, we apply a threshold to identify the hotspot locations.

We compared our method to an exhaustive search strategy and a k-means search strategy. For each sample in the test set, exhaustive search estimates the probability of observing $v^\star$ by looking up the sample in the training set with the most similar distribution to the observed data. This represents the strategy which makes the most use of all the data available for every test sample, at the cost of a linear computational complexity in the number of sample locations in the dataset. In the k-means strategy, we fix a constant test-time complexity by reducing the search space to the $K$ centroids returned by a standard k-means clustering implementation. These centroids are defined such that if each class distribution in the training set were replaced by the nearest of the $K$ centroids, the sum of squared error is approximately minimized, however it does not take into account the sparsity or spatial smoothness of the underlying distributions.

\begin{figure*}
    \centering
    \resizebox{0.9\textwidth}{!}{%
        \begin{tabular}{c}
        \includegraphics{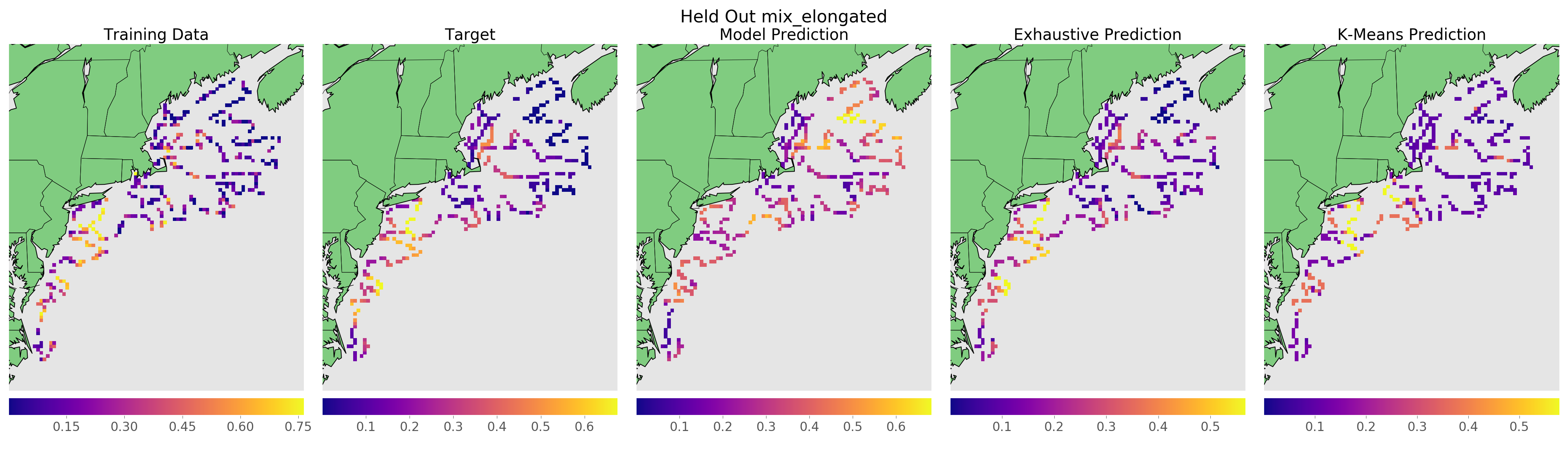} \\
        \includegraphics{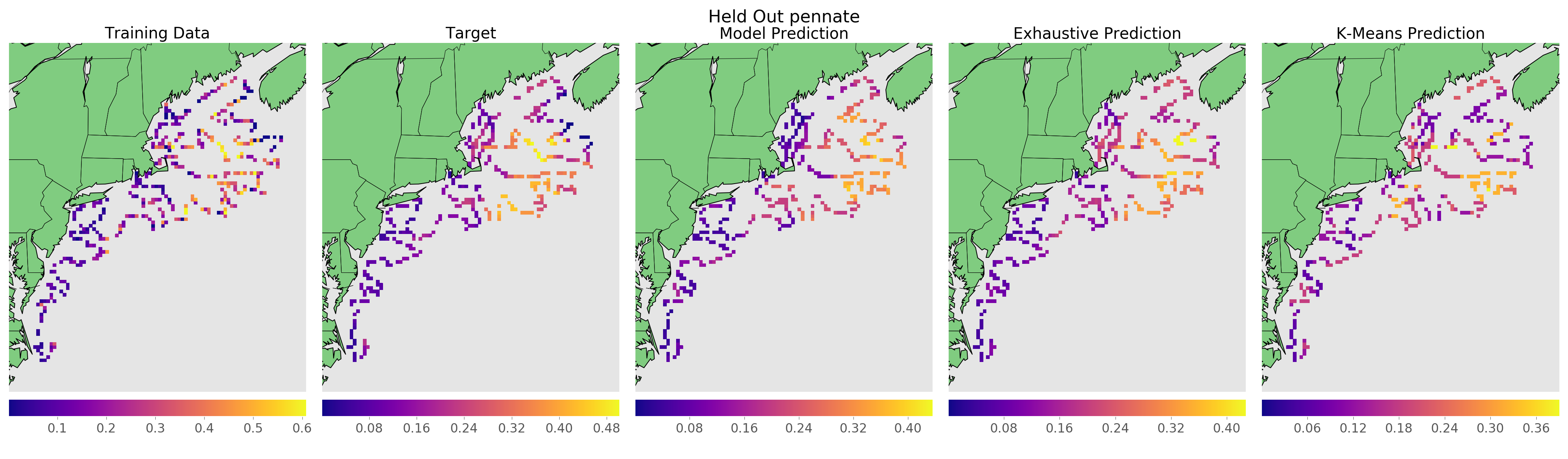}\\
        +
        \includegraphics{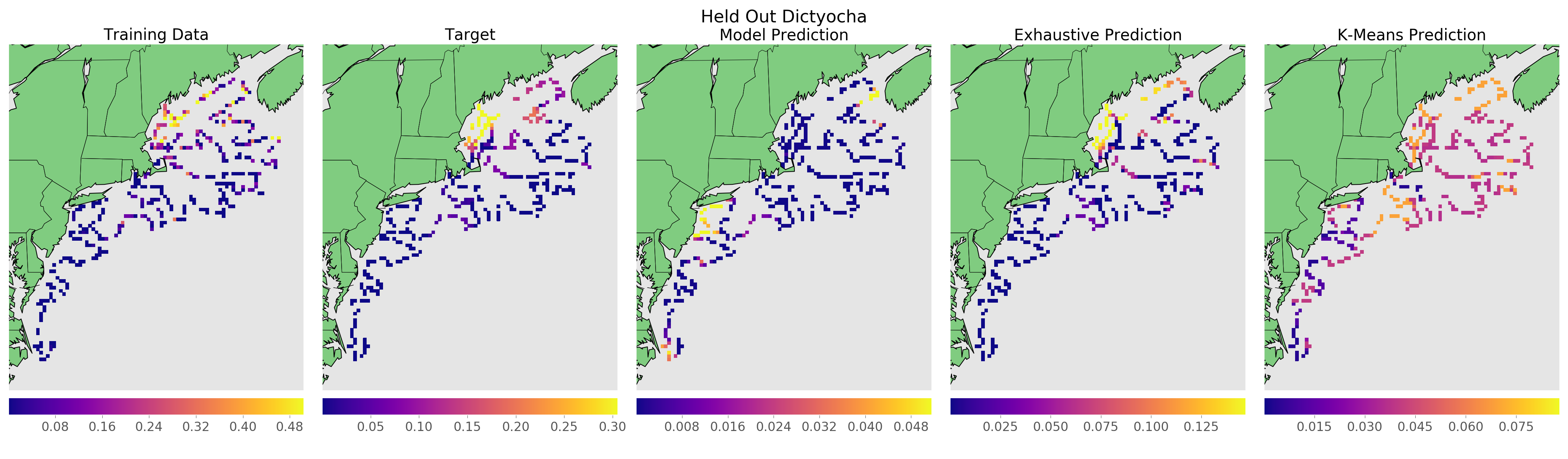}\\
        \includegraphics{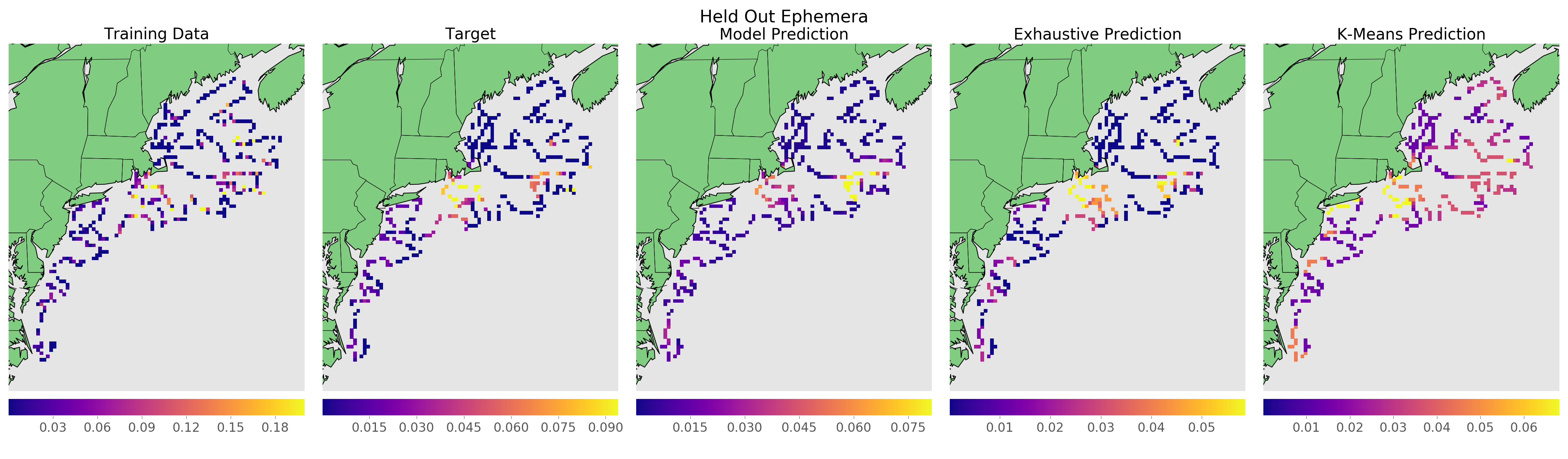}\\
        \end{tabular}%
    }
    \caption{Spatial distribution for four target classes (rows) in interleaved training/testing samples. The columns correspond to training data (col. 1), held-out target locations (col. 2), and the three models under evaluation (col. 3-5). We find close correspondence between the proposed model and the target data, but exhaustive nearest-neighbor approach has the most similar distribution to held-out target locations. This is because the distribution of plankton is correlated with its spatial neighbors, and hence simple interpolation of the training data is likely to give an accurate plankton distribution at the held-out locations.}
    \label{fig:maps_8}
\end{figure*}

\begin{figure*}
    \centering
    \resizebox{0.9\textwidth}{!}{%
        \begin{tabular}{c}
        \includegraphics{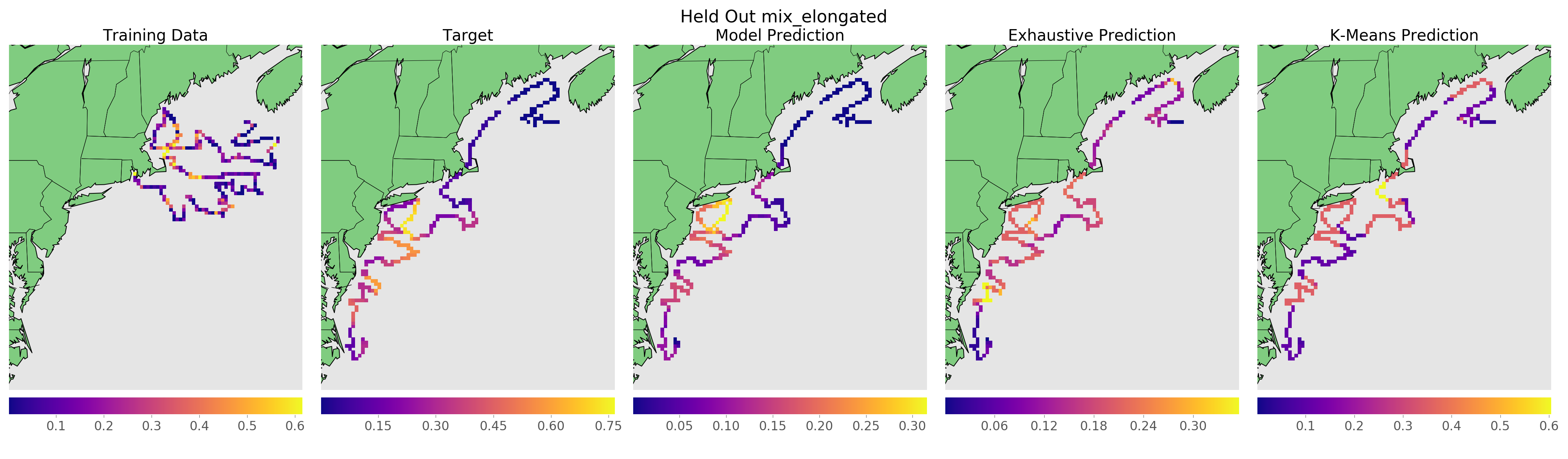} \\
        \includegraphics{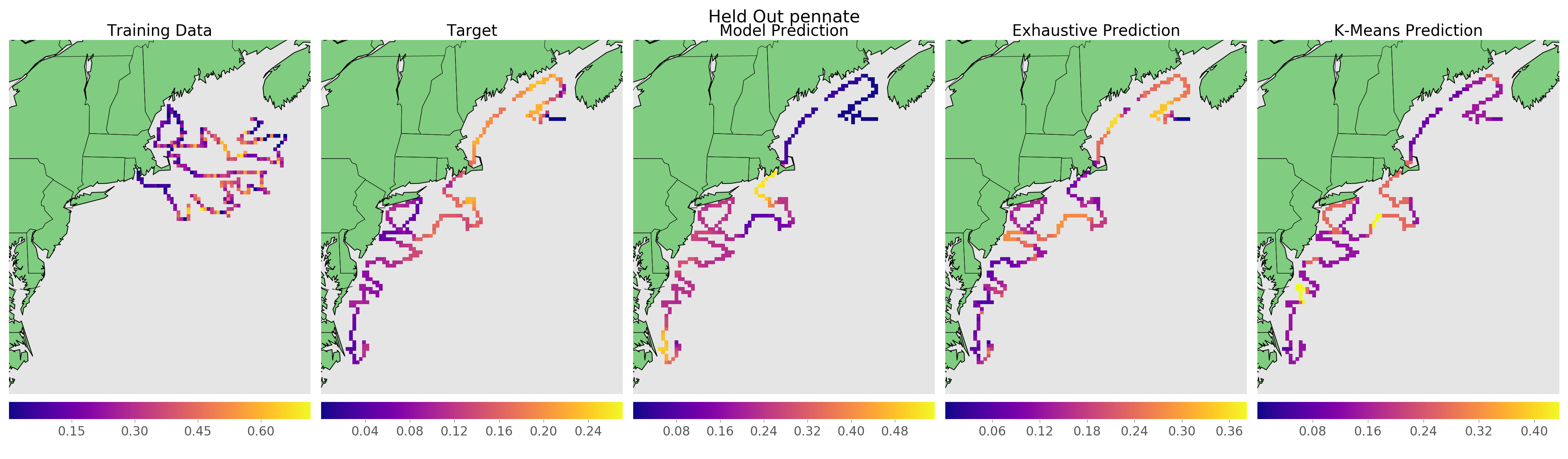}\\
        \includegraphics{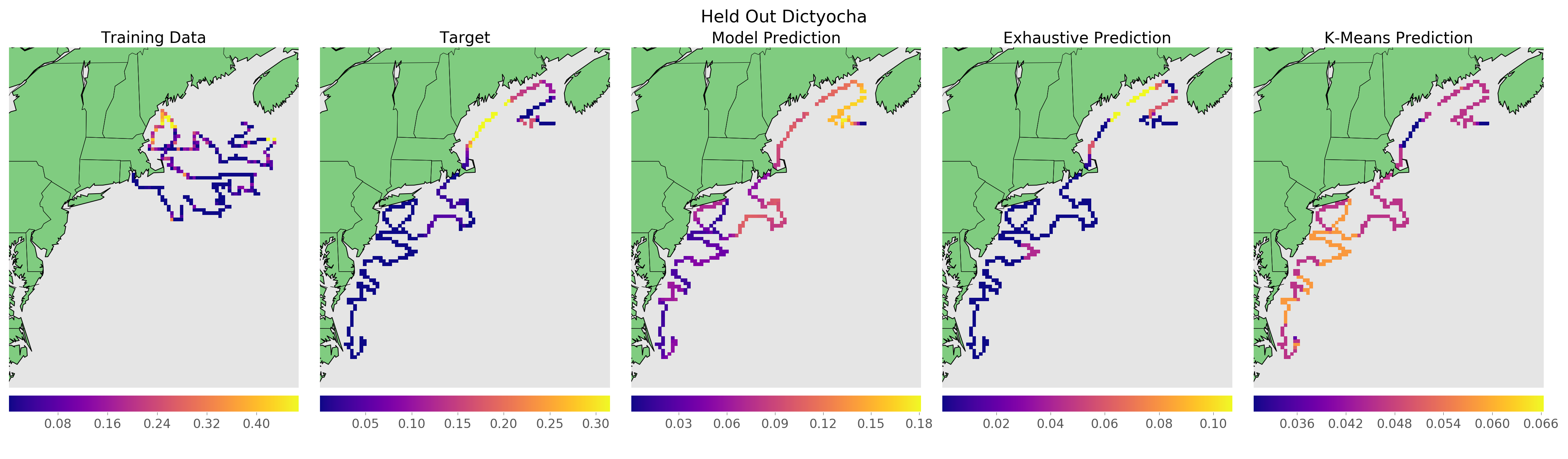}\\
        \includegraphics{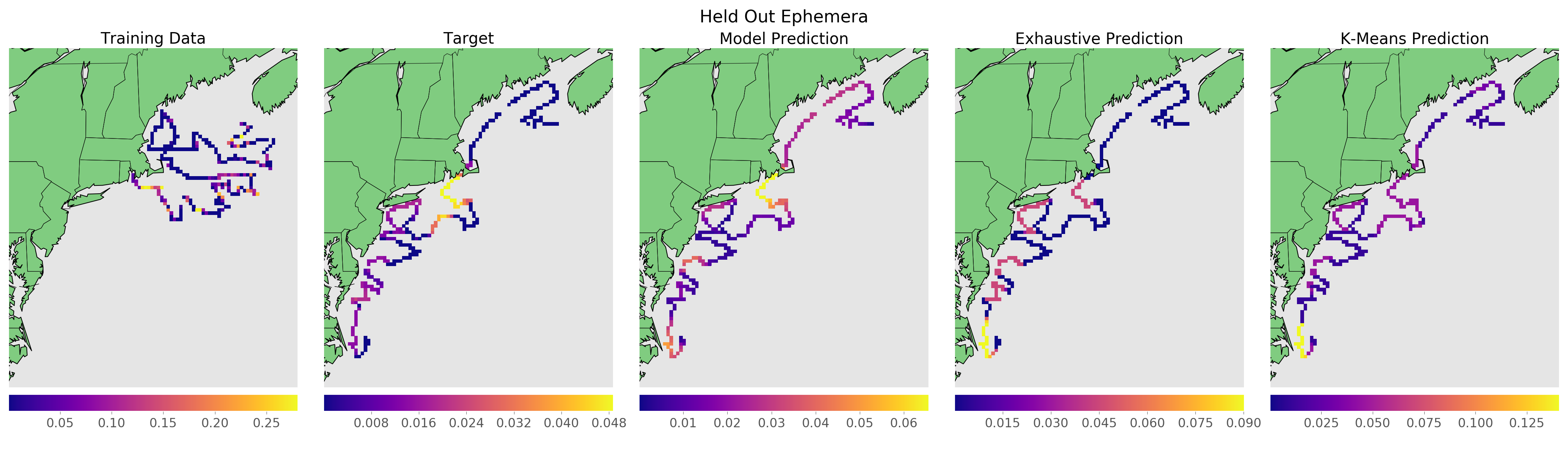}\\
        \end{tabular}%
    }
    \caption{Spatial distribution for four target classes (rows) in split training/testing samples. The columns correspond to training data (col. 1), held-out target locations (col. 2) , and the three models under evaluation (col. 3-5). The proposed plankton topic model provides predictions that agree better with the held-out observations than do the simpler k-means based plankton community model or the exhaustive nearest neighbor search.}
    \label{fig:maps_1}
\end{figure*}

We carried out experiments for two different train/test regimes. First, we used every second sample location for training (see Fig.~\ref{fig:maps_8}, column 1). This regime simulates a mission where the classifier frequently fails to identify examples of a class, for instance because its acceptance threshold was poorly tuned. Because nearby sample locations tend to have similar distributions, this regime tests the ability of a model to interpolate over small distances. Second, we used the first half of the sample locations as training (Fig.~\ref{fig:maps_1}, column 1), and the second half for testing. This latter case simulates a mission where the capabilities of the classifier have changed from the first half to the second half. It tests the ability of a model to predict in a new location that is not likely to have any spatially linked correlation with the training data.

\begin{figure*}[t]
    \centering
    \begin{subfigure}{0.45\textwidth}
        \centering
        \includegraphics[width=1.0\textwidth]{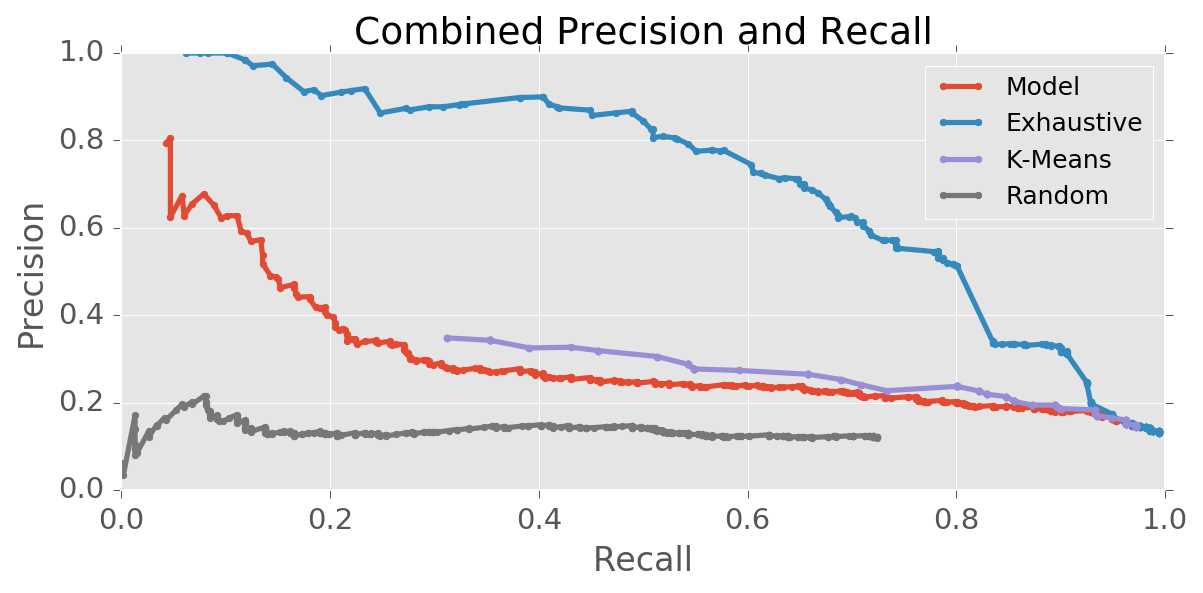}
        \caption{Combined precision-recall, interleaved samples}
        \label{fig:prs_8}
    \end{subfigure}%
    \begin{subfigure}{0.45\textwidth}
        \centering
        \includegraphics[width=1.0\textwidth]{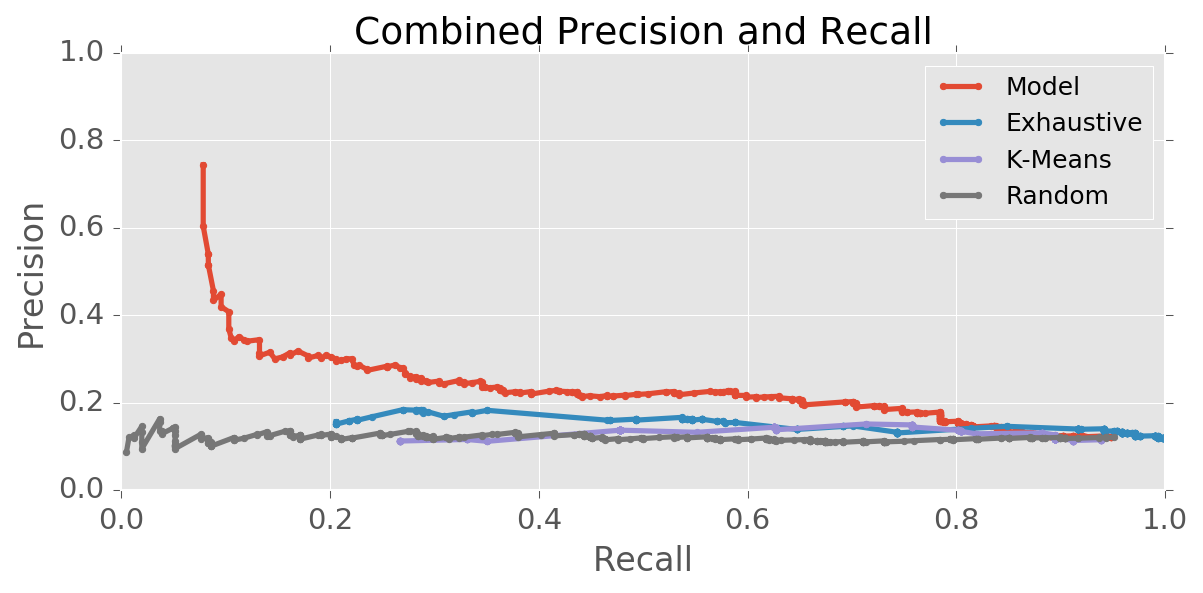}
        \caption{Combined precision-recall, split samples}
        \label{fig:prs_1}
    \end{subfigure} %
    
    \vspace{12pt}
    
    \begin{subfigure}{1.0\textwidth}
        \centering
        \includegraphics[width=1.0\textwidth]{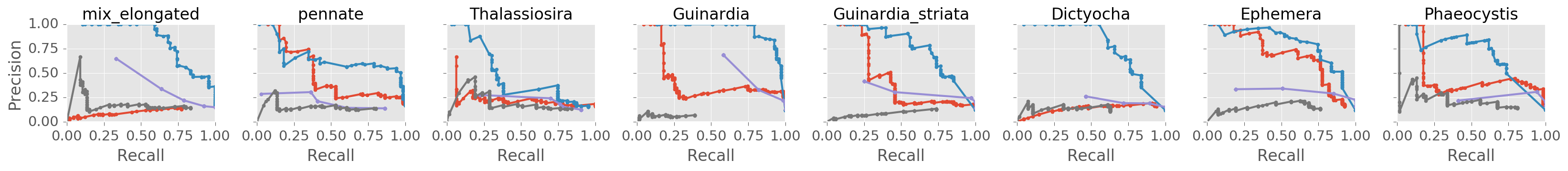}
        \caption{Interleaved samples}
        \label{fig:pr_8}
    \end{subfigure} \\
    \begin{subfigure}{1.0\textwidth}
        \centering
        \includegraphics[width=1.0\textwidth]{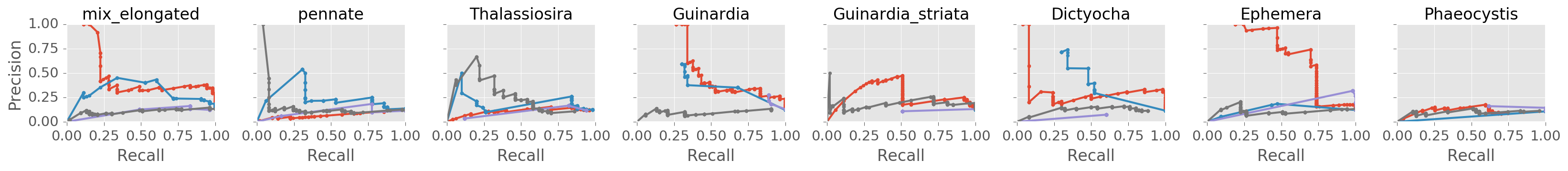}
        \caption{Split samples}
        \label{fig:pr_1}
    \end{subfigure}
    \caption{Plots showing precision-recall curves that indicate the performance of the proposed technique at predicting hotspots of the target plankton species. (a,c) When the training data is interleaved with the target locations, the exhaustive nearest-neighbor has the best average performance. (b,d) The proposed model has the best average performance in cases where observations from nearby locations are not available (split samples), and hence a robust plankton community model is required. }
    \label{fig:pr_all}
\end{figure*}

\section{Results}

We ran our model for a range of choices of the hyperparameters $\alpha \in \{0.001, 0.01, 0.1, 0.5, 1\}$, $\beta \in \{0.001, 0.01, 0.1, 0.5, 1\}$, and $\gamma \in \{10^{-6},10^{-5},10^{-4}\}$ with each of the top 8 taxa held out of the testing data and for both training regimes. We also ran the exhaustive search and k-means strategies for each. The strategies each produce an estimate for $P(w=v^\star | c)$, which we then smooth with a median filter with size parameter $\sigma$. For a scalar threshold $\tau$, we predict that cell $c$ is a hotspot if $\Pi_\sigma(P(w=v^\star | c)) > \tau$, where $\Pi_\sigma$ is the median function over a square region with side length $\sigma$. 

To evaluate our results we compare the held-out locations in the test set (Fig.~\ref{fig:maps_8} and \ref{fig:maps_1}, column 2) to predictions from each of the proposed strategies (Fig.~\ref{fig:maps_8} and \ref{fig:maps_1}, our model, column 3; exhaustive nearest neighbour search, column 4; and k-means search, column 5). The input to the models is illustrated with the observed values of the held-out class at the training locations (Fig.~\ref{fig:maps_8} and \ref{fig:maps_1}, column 1). Our findings show that the prediction problem is relatively straightforward for the interleaved experiment (Fig.~\ref{fig:maps_8}). In contrast, the problem is much more difficult when training and testing locations are in different parts of the world. (Fig.~\ref{fig:maps_1}). Despite this, for three (Fig.~\ref{fig:maps_8} and \ref{fig:maps_1}, rows 1, 2, 4) of the four target classes shown here, the spatial location of maxima of our model's predictions are consistently near the maxima in the target distributions.

Varying $\tau$ for each strategy and parameter setting we can count the true positive, true negative, false positive, and false negative hotspot predictions compared to the top 50 examples in the held-out data. These counts give the precision and recall for each parameter choice, for each $v^\star$. We also accumulate these counts across all $v^\star$ to compute the overall precision and recall for each choice of parameters. We assign each set of parameters a score given by the area under its aggregated precision-recall curve and select the parameter set with the maximum score for further comparisons. For the interleaved experiment, best performance was achieved with $\alpha=0.1, \beta=0.1, \gamma=10^{-5}, \sigma=25 km$ and for the split experiment, $\alpha=0.1, \beta=1.0, \gamma=10^{-5},\sigma=35 km$. We chose the number of centroids for the k-means strategy to be the same as the number of topics in the best performing topic model, $K=9$ for the interleaved experiment, and $K=6$ for the split experiment.

We compare the aggregated and individual class precision-recall curves for the best parameters for each strategy (Fig.~\ref{fig:pr_all}). Note that precision refers to the ratio of the number of correctly predicted hotspots to the total number of predicted hotspots, and that recall refers to the ratio of correctly predicted hotspots to the total number of real hotspots. An ideal algorithm will have precision of 1 and recall of 1.  From the aggregated precision-recall curves, we find that our model significantly outperforms the exhaustive nearest-neighbor and the k-means strategies on the split-samples regime, especially for low recall requirements. This indicates that the top few predictions of our model were more likely to be true hotspots than those of the other strategies. The exhaustive nearest-neighbor strategy barely performs better than random guessing on the split regime, yet it performs extremely well on the interleaved regime. This result is expected as the exhaustive strategy does not reason at all about the underlying association between plankton types. Instead, it depends on having observed a training point whose distribution is similar to every test point. In contrast, our model performs nearly as well on the split regime as the interleaved regime.

Our model also outperforms the k-means strategy on the split-samples regime. Note that the k-means strategy is exactly equivalent to the exhaustive search strategy in the limit where $K$ is the number of training points. Both these strategies rely on a distance metric over the class distributions. The high dimensionality of the distributions acts to the detriment of the distance metric. As the dimensionality of a space increases, the discriminating power of distance metrics within that space decreases. The amount of data needed to find meaningful clusters grows exponentially with the number of dimensions, a phenomenon sometimes called \textit{the curse of dimensionality}. As a result, the two search-based strategies perform well when test points are very near to training points in taxon distribution space, but when test points are further away, a distance metric is less informative and performance is negatively impacted. Our model mitigates this problem with additional constraints in the form of a hierarchical generative model, the CRP prior on the spatial distribution of communities, and sparse Dirichlet priors on the plankton class distribution that describes each community.

\section{Discussion}
Our ongoing efforts are focused on using this work to improve on autonomous sampling techniques for sparsely distributed class counts, including phytoplankton taxa. In this work we have considered the case where the classifier fails to make any predictions whatsoever about some class, however the results are also relevant to scenarios where unexpectedly low or high numbers of a class are observed. Recall that IFCB samples only \SI{5}{\milli\liter} of water at a time, yet researchers would like to characterize the plankton distribution in a wide area of ocean. As a result, the measured distributions in individual cells are extremely noisy. This was not taken into account while collecting the present dataset, and as a result we need to use spatial smoothing on the order of a 15km radius to achieve meaningful predictions.
An interesting future direction for this work is to compare our model's prediction to real-time measurements, and use this comparison to decide whether more data is needed. We are currently developing a variant of IFCB that can be deployed on autonomous surface vehicles such as the WHOI JetYak which will allow dynamic planning with respect to the plankton observations and our model.

We also plan to explore further models which build on the one proposed in this work. A natural research direction is to develop models of the relationships between taxa and environment variables. However, the high-dimensionality, sparsity of observations, and noisiness of the distribution of individual taxa make learning these relationships difficult. Our initial explorations have suggested that the relationships between environment variables and communities are easier to characterize with simple models than those with individual taxa. We are particularly interested in such models which also incorporate temporal aspects, as they could enable learning causal relationships involved in phytoplankton lifecycles and the changing ecosystems. Finally, we plan to explore deeper generative models of the observations, which we expect will discover more complex community structures.

\section{Conclusion}
We have presented a novel technique for finding hotspots of discrete targets that are sparsely distributed in the world. The proposed method utilizes a probabilistic generative model to describe spatial co-occurrence relationships between the target and other kinds of observations. Our technique uses these relationships to estimate the target's spatial distribution in locations where robust measurements are not available. We apply our approach to the problem of finding hotspots of phytoplankton taxa in observations made by a robotic marine instrument.

The proposed technique utilizes a mixture model with spatial smoothness and sparsity constraints on phytoplankton distributions to enable accurate predictions, even when the observed plankton distribution is very different from training data. We validated our approach with real data collected on a two week fixed-trajectory survey mission. Results from experiments show that our model produces a better community representation that can more accurately predict hotspot locations than either exhaustive nearest-neighbour search or a k-means based plankton community model.

\section*{Acknowledgement}
\small{This work was supported in part by awards to YG from NOAA through its Cooperative Institute for the North Atlantic Region (CINAR) program, and  from WHOI; and to HMS from NASA's Ocean Biology and Biogeochemistry Program, and from NOAA through CINAR. We are indebted to Emily Brownlee for expert assistance with IFCB data collection and Joe Futrelle for facilitating IFCB data access and analysis workflows. We also thank the captain and crew of the Research Vessel Pisces and scientists from NOAA's Northeast Fisheries Science Center for enabling our participation in EcoMon surveys. We gratefully acknowledge the support via grant to GD of the Natural Sciences and Engineering Research Council of Canada (NSERC).}

\bibliographystyle{IEEEtran}
\bibliography{IEEEabrv,girdhar,arnold}
%%%%%%%%%%%%%%%%%%%%%%%%%%%%%%%%%%%%%%%%%%%%%%%%%%%%%%%%%%%%%%%%%%%%%%%%%%%%%%%%

\end{document}